\documentclass{article}

\usepackage[preprint]{neurips_2022}

\usepackage[utf8]{inputenc} 
\usepackage[T1]{fontenc}    
\usepackage{hyperref}       
\usepackage{url}            
\usepackage{booktabs}       
\usepackage{amsfonts}       
\usepackage{nicefrac}       
\usepackage{microtype}      
\usepackage{xcolor}         

\setcitestyle{numbers,square}
\usepackage{multirow}
\usepackage{amsmath}
\usepackage{caption}
\usepackage{subcaption}
\usepackage{graphicx}
\title{Sports Re-ID: Improving Re-Identification Of Players In Broadcast Videos Of Team Sports}

\author{%
  Bharath Comandur \\
  Apple Inc.\\
  \texttt{cjrbharath@gmail.com} \\
}

\begin{document}

\maketitle

\begin{abstract}
  Re-identification (re-id) of people in images is a well-studied problem in computer vision for many applications. This work focuses on player re-identification in broadcast videos of team sports. Specifically, we focus on identifying the same player in images captured from different camera viewpoints during any given moment of a match. This task differs from traditional applications of person re-id in a few important ways. Firstly, players from the same team wear highly similar clothes, such as a team jersey/uniform, thereby making it harder to tell them apart. Secondly, there are only a few number of samples for each identity, which makes it harder to train a re-id system. Thirdly, the resolutions of the images are often quite low and vary a lot. This combined with heavy occlusions and fast movements of players greatly increase the challenges for re-id. In this paper, we propose a simple but effective hierarchical data sampling procedure and a centroid loss function that, when used together, increase the mean average precision (mAP) by 7 - 11.5 and the rank-1 (R1) by 8.8 - 14.9 without any change in the network or hyper-parameters used. Our data sampling procedure improves the similarity of the training and test distributions, and thereby aids in creating better estimates of the centroids of the embeddings (or feature vectors). Surprisingly, our study shows that in the presence of severely limited data, as is the case for our application, a simple centroid loss function based on euclidean distances significantly outperforms the popular triplet-centroid loss function. Our proposals provide comparable improvements for both convolutional networks and vision transformers. Our approach is currently ranked \#1 on the ongoing SoccerNet Re-Identification Challenge 2022 leaderboard (test-split) with a mAP of 86.0 and a R1 of 81.5. On the sequestered challenge split, we achieve an mAP of 84.9 and a R1 of 80.1. While we demonstrate results on soccer matches, our proposals naturally extend to any team sport. Research on re-id for sports-related applications is very limited and our work presents one of the first discussions in the literature on this. Code is available at \href{https://github.com/shallowlearn/sportsreid}{https://github.com/shallowlearn/sportsreid}.
 
\end{abstract}

\section{Introduction}
\label{sec:introduction}

\begin{figure}[h]
     \begin{subfigure}{0.48\textwidth}
         \centering
         \includegraphics[width=1\textwidth,height=0.5\textwidth]{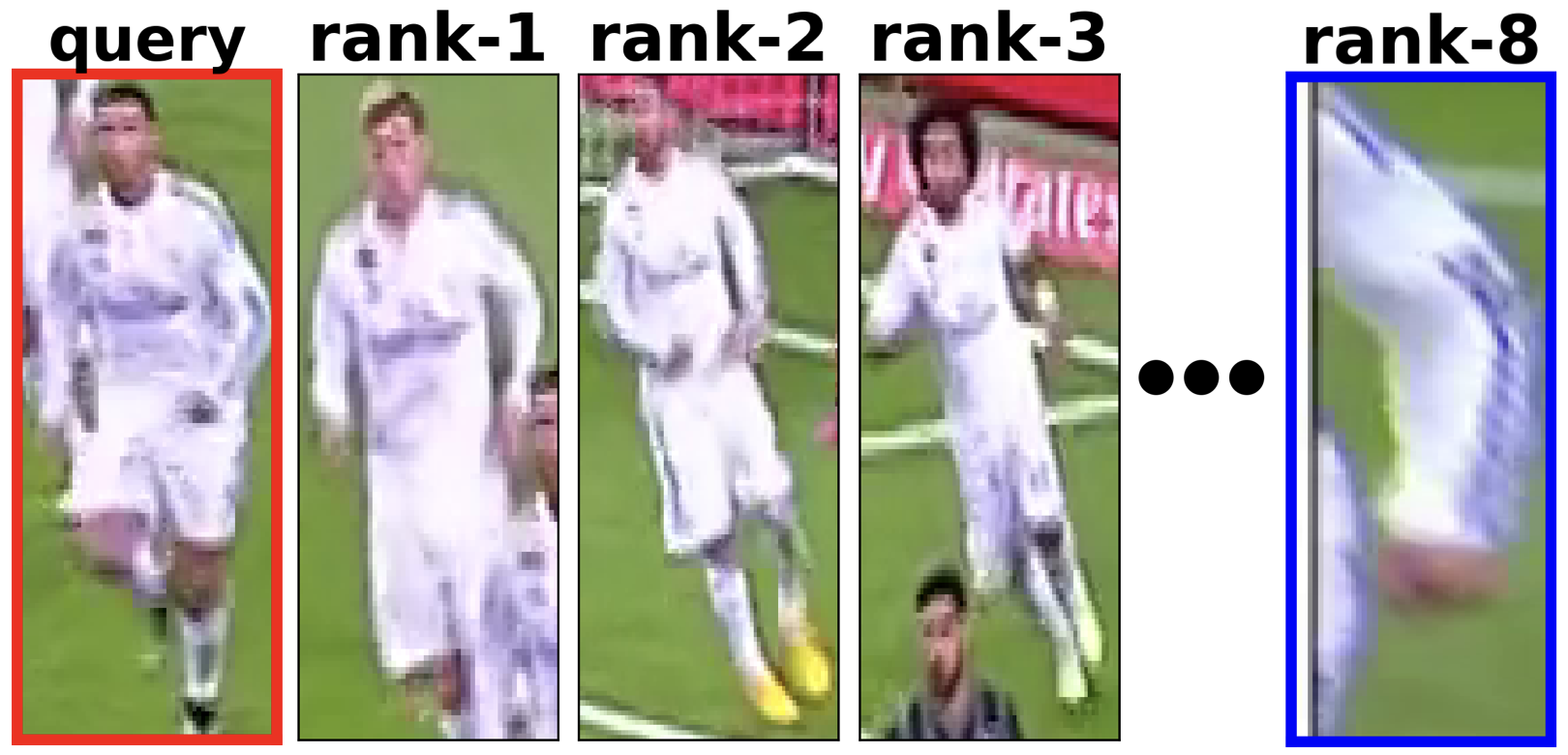}
         \caption{}
         \label{fig:a}
     \end{subfigure}
     \hfill
    \begin{subfigure}{0.48\textwidth}
        \centering
        \includegraphics[width=1\textwidth,height=0.5\textwidth]{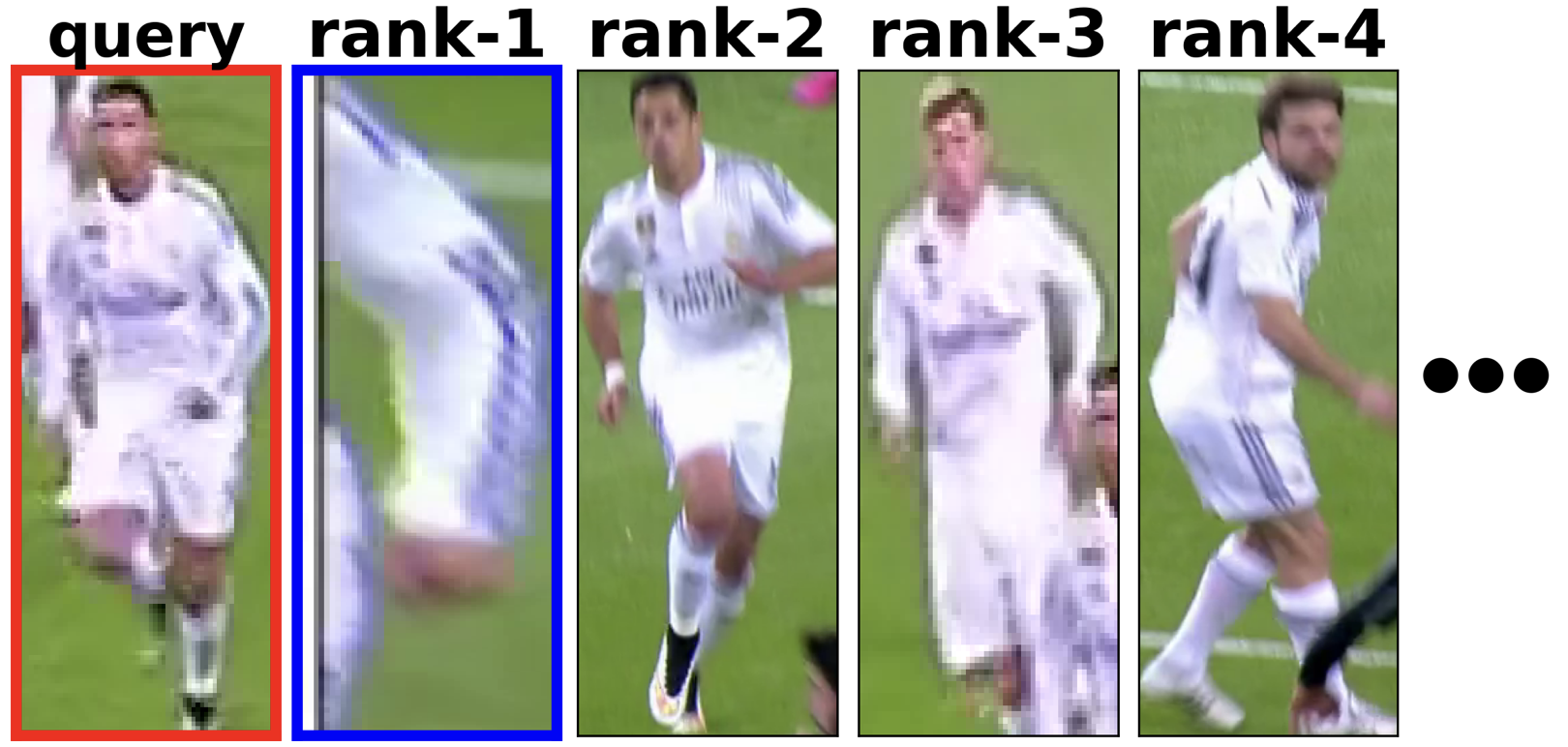}
        \caption{}
        \label{fig:b}
    \end{subfigure}
     \begin{subfigure}{0.48\textwidth}
         \centering
         \includegraphics[width=\textwidth,height=0.5\textwidth]{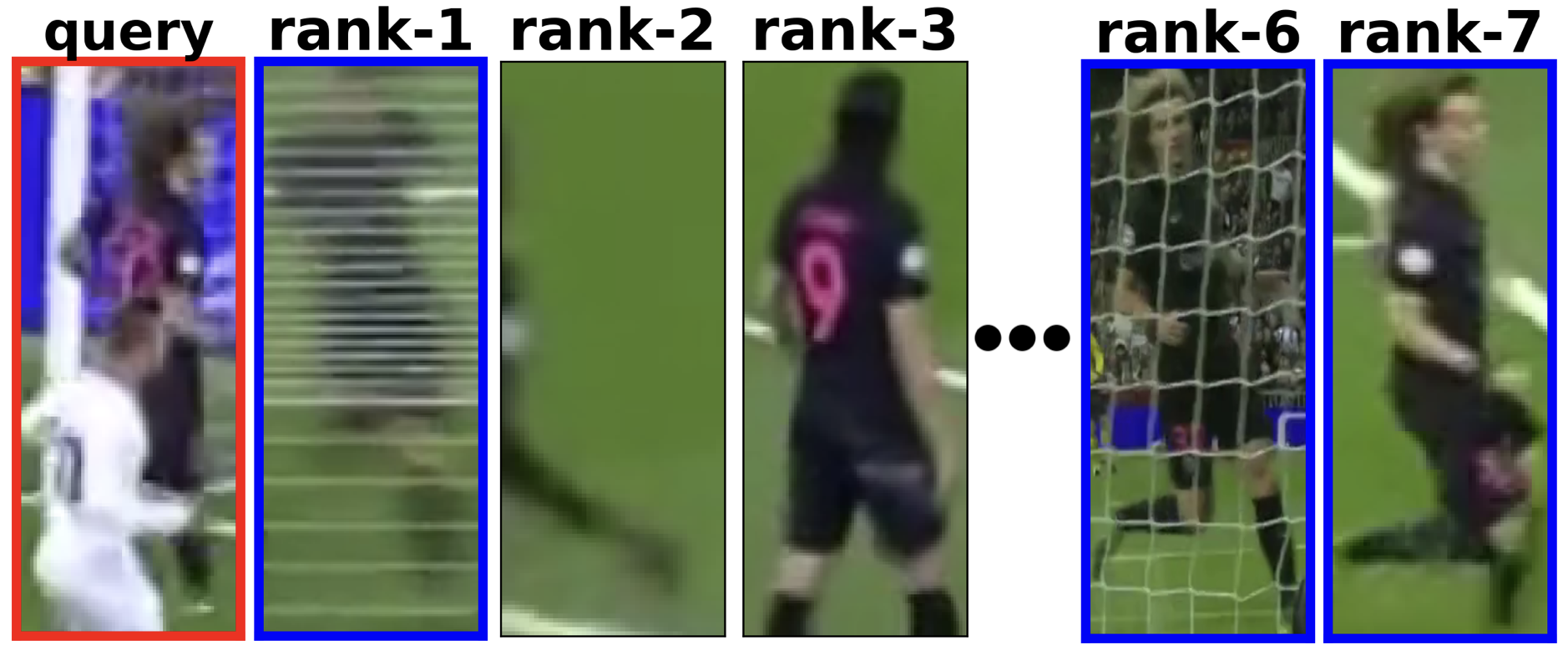}
         \caption{}
         \label{fig:c}
     \end{subfigure}
     \hfill
     \begin{subfigure}{0.48\textwidth}
         \centering
         \includegraphics[width=\textwidth,height=0.5\textwidth]{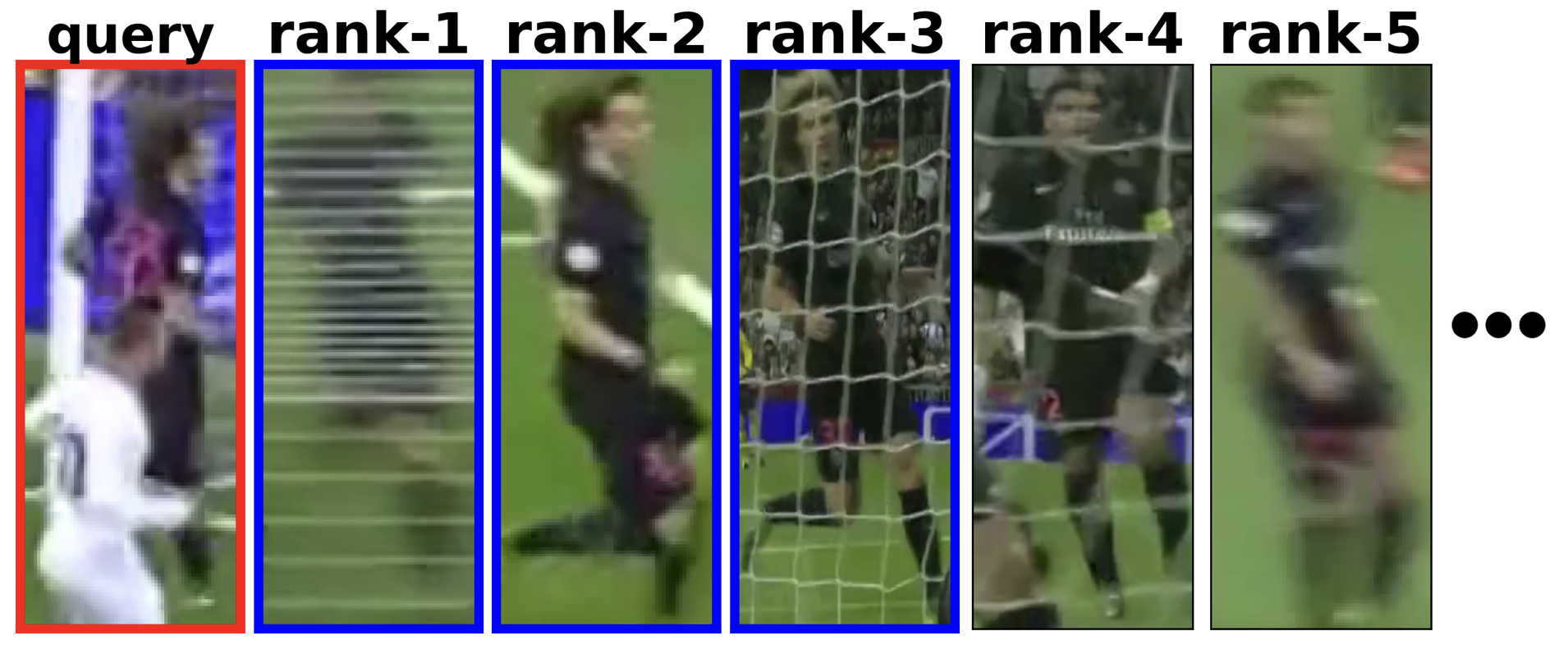}
         \caption{}
         \label{fig:d}
     \end{subfigure}
    \caption{Some examples of the power of our approach in challenging scenarios. The results in the column on the left are obtained by training an OSNet \cite{zhou2019omni} using random data sampling and triplet loss \cite{hermans2017defense}. The results in the column on the right are obtained by training the same network with our proposed hierarchical sampling and centroid loss. The query image is highlighted with a \textcolor{red}{red} border and the true matches in the corresponding gallery images are marked with a \textcolor{blue}{blue} border. The gallery images are ranked in increasing order of their predicted distances from the query. Comparing Figs.~\ref{fig:a} and \ref{fig:b}, we see that our procedure helps the same network learn powerful features that can correctly re-identify a player even if the gallery image contains only a small part of the player in the query image. Comparing Fig.~\ref{fig:c} with Fig.~\ref{fig:d} shows that our approach correctly identifies a player from different viewpoints even in the presence of occlusions.}
    \label{fig:results}
\end{figure}

Person re-identification (re-id) in images and videos is quite useful for many applications that require tracking people across multiple cameras. Examples of such applications include surveillance in airports and public spaces, tracking customers in automated grocery stores such as Amazon Go, etc. While there exists a lot of prior art on person re-id for surveillance, in this paper, we focus on an interesting application of re-id that is under-represented in the literature. Our application involves re-identifying players in broadcast videos of team sports such as soccer, basketball, etc. Specifically, we want to develop a system that can quickly re-identify players across different camera viewpoints at any given moment in a match. Such a system can be used for many applications including tracking players across multiple cameras, building automatic highlight videos that focus on a single player, developing better tools to assist referees, etc. In fact, this is the main objective of the first edition of the SoccerNet Re-Identification Challenge 2022 \cite{soccerNetv3,soccerNetv2}.

This re-id task differs from traditional re-id applications, such as for surveillance, in the following important ways -- 1) players that belong to the same team often look very similar because they wear similar uniforms or team jerseys; 2) the resolutions of the images are often quite low and vary a lot. This combined with significant occlusions, and fast movements of players greatly increase the challenges for re-id. Figs. \ref{fig:results} and \ref{fig:addfigures} illustrate these issues; 3) there are only a few number of samples for each identity, which makes it harder to train machine-learning models.

Through extensive evaluation in Section \ref{sec:results}, our study shows that due to these differences, the performance of re-id systems that have been developed for surveillance applications, degrades when applied to the task of player re-id without any modifications.



Our contributions are summarized as follows:
\begin{enumerate}
    \item We present one of the first research studies that is specifically focused on re-identifying players in broadcast videos of team sports.
    \item Via extensive evaluation, we show that conventional data sampling and batching strategies used for surveillance re-id are not suitable for our application, as they create dissimilar data distributions during training and testing.
    \item Motivated by the above observation, we propose a simple but effective hierarchical data sampling and batching procedure to reduce the gap between the training and test data distributions. 
    \item Many SOTA approaches for re-id such as \cite{wieczorek2021unreasonable, do2019theoretically, yuan2020defense, zhang2020beyond, wang2019centroid, lagunes2020centroids, alnissany2022modified} use a triplet-centroid loss to improve upon the popular triplet loss \cite{hermans2017defense}. However, our study shows that in the presence of severely limited data, as is the case for our application, a triplet-centroid loss hardly offers any improvement over a triplet loss. Surprisingly, we show that a simple centroid loss function based on euclidean distances significantly outperforms triplet-centroid loss functions for our application.
    \item Our proposals yield an increase in the range of 7 - 11.5 for the mean average precision (mAP) and an increase in the range of 8.8 - 14.9 for the rank-1 (R1), without changing the network or hyper-parameters. We demonstrate comparable increases in mAP and R1 for both convolutional neural networks and vision transformers \cite{dosovitskiy2020image, touvron2021training}.
\end{enumerate}

\section{Related work}
\label{sec:related_work}

An exhaustive survey of prior art in re-id is beyond the scope of this work, since person re-id is an extremely popular research area. The studies in \cite{wu2019deep, wang2019beyond, radhliterature} provide an in-depth survey of existing re-id techniques. On the one hand, the algorithms in \cite{yuan2020defense, do2019theoretically, yuan2020defense, zhang2020beyond, wang2019centroid, lagunes2020centroids, wieczorek2021unreasonable, alnissany2022modified} represent a whole category of approaches that focus on developing suitable loss functions to improve re-id. On the other hand, the work in \cite{he2021transreid, zhou2019omni, zhou2021learning} represent a category of approaches that concentrate on developing suitable networks to extract multi-scale, multi-view invariant features. The authors of \cite{he2021transreid} apply transformers to re-id while those of \cite{zhou2019omni, zhou2021learning} develop 'omni-scale' features for finegrained re-id.

Other techniques including the ones in \cite{groupreid1,groupreid2,groupreid3,groupreid4,groupreid5,groupreid6} propose hierarchical clustering and grouping strategies for re-id. However, they use the term 'hierarchical' in the context of feature-based grouping or clustering, and subsequently learning a mapping function to assign each person to a position within the group structure. This is different from our context, where we use a simple, effective and deterministic rule-based approach for hierarchically sampling data. Our approach is suitable for many team sports.

Some popular public datasets for re-id include CUHK03\cite{CUHK03}, Market-1501 \cite{Market-1501}, MSMT17 \cite{MSMT17}, Street2Shop\cite{Street2Shop}, etc. However, almost all of them are for surveillance or fashion related applications. To the best of our knowledge, the most relevant public sports-focused re-id contribution that we found is the work in \cite{penate2020tgc20reid} that creates a dataset for re-id in long-distance running. Research on re-id in sports is very limited. Infact, this year is the first edition of the SoccerNet \cite{soccerNetv3, soccerNetv2} Re-ID Challenge. There is a lot of avenues for new research in re-id for sports applications.

\section{Some basic concepts and definitions for person re-identification}
\label{sec:background}

Consider an image that contains a person of interest. We refer to this image as the query image. We want to identify this person in another collection of images that we refer to as the gallery images. Broadly speaking, many SOTA approaches \cite{wieczorek2021unreasonable, he2021transreid, zhou2021learning} that employ neural networks for re-id, use some variant of the following procedure. 

Each unique person in the training data is assigned a unique id. An object detector is used to detect bounding boxes of each person in an image. The image is then cropped using these bounding boxes to yield a set of detections per image. The class label for each such detection is the id of the person that it contains. Let $Q$ denote a detection produced from a query image that contains the person of interest. We refer to this as the query detection. Let $G = \cup G_i$ denote the set of all detections obtained from the gallery images. We refer to $G$ as gallery detections. $G_i$ is the $i^{th}$ gallery detection. These detections are then fed to a classifier such as a neural network. This network is trained using a combination of loss functions such as a classification (cross-entropy) loss and a triplet loss \cite{hermans2017defense}. At inference, the final layer of the network that outputs the class scores is discarded. For each input detection, the output of the penultimate layer of the network is used as its corresponding feature vector or embedding.  Let $F_Q$ denote the embedding of $Q$, and $F_{G_i}$ denote the embedding of $G_i$. $F_G = \cup F_{G_i} $. The distance between $F_Q$ and every element $F_{G_i}$ $\epsilon$ $F_G$ is computed. This distance can be an euclidean distance or cosine distance, etc. For some applications, the $G_i$ that has the smallest distance to $Q$ is selected as the match. For other applications, the $G_i$'s are ranked in increasing order of their distances from $Q$. 

\subsection{Random sampling, triplet loss, and BATCH HARD}
\label{sec:triplet_batchhard}
The novel study in \cite{hermans2017defense} shows that the procedure used to create training batches is key when using a triplet loss for re-id. They propose the following approach that they refer to as BATCH HARD. 

For each batch, we collect $K$ samples for a specific id. We do this for $M$ randomly sampled unique ids, resulting in a batch size of $K \cdot M$. We apply the network on this batch to predict the class label and the embedding for each sample. We then compute all pairwise distances between the embeddings in this batch. Let $D_{I,J}$ denote the distance between two detections I and J. Since the $M$ ids in this batch are randomly sampled, we refer to this procedure as random sampling. We treat each sample in this batch as a query detection, and treat the other samples in the batch as the corresponding gallery detections. Thus for each sample $A$, we find the hardest positive $P$ within the batch, i.e., the sample that is farthest from $A$ while having the same id as $A$. We also find the hardest negative $N$ within the batch, i.e., the sample that is nearest to $A$ but has a different id than $A$. Nearest and farthest are calculated using the pairwise distances. 

The triplet loss for this sample is then defined as

\begin{equation}
    L_{T}(A) = [m + D_{A,P} - D_{A,N}]_{+}
\end{equation}
where $[W]_+$ denotes $\text{max}(0, W)$. $P$ and $A$ have the same id. $N$ and $A$ have different ids.

The triplet loss for a batch $B$ is simply the sum over the samples.
\begin{equation}
\label{eq:tripletloss}
    L_{T} = \sum_{A, A \epsilon B} [m + D_{A,P} - D_{A,N}]_{+}
\end{equation}



\subsection{Player re-id in broadcast videos of team sports}
\label{sec:sports-reid}
Consider a broadcast video of a basketball game or soccer match. We refer to each moment of interest in the match as an "action". We call the first frame in which we see an action in the broadcast video as the "action frame" for that action. A broadcast video often contains replays captured from other camera viewpoints. This means that the same action can appear in the broadcast video at later frames. We call the other frames that depict the same action, and that appear later in the broadcast video, as "replay frames". Thus while an action frame and its corresponding replay frames appear at different times in the broadcast video, they represent roughly the same timestamp with respect to the match itself. For example, an action occurring at the 5 minute mark in a match can have the action frame at frame number 100 and a replay frame at frame number 1000.

Our objective is to re-id a player in the action frame, across the different replay frames for that action. This is highly useful for tracking players across multiple cameras, creating highlight videos, virtual assistance for referees, etc. Note that person detection is beyond the scope of this paper. We assume that the we already have the person detections for the action and replay frames. Note that due to the nature of team sports, a detection can contain multiple players with occlusions as shown in Figs. \ref{fig:results} and \ref{fig:addfigures}. This makes re-id challenging. In our context, each detection from a given action frame is a query and the detections from the corresponding replay frames are the gallery detections for that query. Our re-id system ranks these replay frames based on their distance to the query.

\section {Our proposals to improve player re-id in broadcast videos}

\subsection{Motivation}

At inference, the people in the gallery and query detections in one batch will belong to one of two teams, or in some cases be a referee. Now consider a single batch during training time, that is created using the BATCH HARD procedure \cite{hermans2017defense} described in Section \ref{sec:triplet_batchhard}. This batch consists of $K$ samples for each of $M$ different ids. Hence the batch size is $K\cdot M$. We now make the following hypothesis. If the $M$ different ids for this single batch are \emph{randomly sampled} from the entire dataset consisting of samples from many matches (which is the conventional approach), then the distribution of the pairwise differences between the samples of this batch will be different from the distribution of the pairwise differences between the samples of the batch that the network sees during inference. For instance, with random sampling of the ids, each training batch contains images of players from multiple teams, and the network potentially learns to extract image features that exploit the differences in the clothing. However, such features won't have sufficient discriminative power if we are comparing a player against other players from the same team, \emph{which is precisely the case at inference time}.


\subsection{Hierarchical data sampling}
Based on the hypothesis described above, we propose an alternate hierarchical data sampling procedure that we denote as hierarchical sampling. To create this hierarchy, we make use of the metadata that comes with the images. For each sample at training, we know its "action" (Section \ref{sec:sports-reid}), which match it belongs to, the year in which the match was played, and also the names of the two teams participating in that match. Note that we do not know which of the two teams the player belongs to. Using this information, we hierarchically group the data into multiple levels as shown in Table \ref{table:hierarchical_levels}: 

\begin{table}
  \caption{Hierarchical grouping for data sampling and batching}
  \label{table:hierarchical_levels}
  \centering
  \begin{tabular}{ll}
    \toprule
    Level     & Description  \\
    \midrule
    O & A random sample with its action, match, year and two participating teams. \\
    I & All samples with the same action. See Section \ref{sec:sports-reid} for the definition of an "action". \\
    II  & All samples from the same match \\
    III  & All samples from any match between the same two teams in the same year \\
    IV & All samples from any match between the same two teams in any year \\
    V & All samples from any match that involves at least one of the two teams in the same year\\
    VI & All samples from any match that involves at least one of the two teams in any year \\
    VII & All samples\\
    \bottomrule
  \end{tabular}
\end{table}

For a single training batch, we randomly select a sample (Level O in Table \ref{table:hierarchical_levels}). We then proceed to create a training batch by first sampling from Level I in Table \ref{table:hierarchical_levels}. If we do not have sufficient samples for a batch, we proceed to Level II, while excluding samples that have already been added to the batch. In this fashion, we proceed from level to level until we have enough samples for a batch. All the samples selected in this batch are then discarded from the pool of available samples (for this epoch). For the next epoch, we re-initialize the pool of available samples, and again apply the same procedure. In this way, we ensure that the batches are randomized. So a single sample is compared against different samples in different epochs. This hierarchical procedure increases the possibility of having similar samples in each batch, which is the case at inference. 

One potential drawback of our proposed sampling procedure is that it does reduce the total number of possible batches that can be created for training, when compared to pure random sampling. In addition, as mentioned before, there are only a few number of samples for each identity, which makes it harder to train a network for our application. We address this via the use of a centroid loss.

\subsection{Centroid loss vs triplet-centroid loss}


\subsubsection{Triple-centroid loss}

The triplet loss in equation \ref{eq:tripletloss} operates on a per-sample level, meaning that we compare a sample embedding $A$ with two other embeddings, $P$ and $N$, where $P$ is the hardest positive for $A$ within a batch and $N$ is the hardest negative for $A$ within a batch. Hence it can be sensitive to outliers. To handle this, triplet loss functions using centroids have been proposed in prior work such as \cite{do2019theoretically, yuan2020defense, zhang2020beyond, wang2019centroid, lagunes2020centroids, wieczorek2021unreasonable, alnissany2022modified}. However, it is important to note that these studies employ a \emph{triplet (or n-tuplet) loss that uses centroids.} For example, the study in \cite{wieczorek2021unreasonable}, that is currently among the SOTA approaches on many public re-id benchmarks, defines a triplet-centroid loss for a sample $A$ as follows:

\begin{equation}
\label{eq:triplet_centroid}
    L_{TC}(A) = [m + D_{A,\text{Centroid}_P} - D_{A, \text{Centroid}_N}]_{+}
\end{equation}
where $[W]_+$ denotes $\text{max}(0, W)$. $\text{Centroid}_P$ denotes the centroid of the cluster with the same id as $A$ and $\text{Centroid}_N$ is the centroid of the cluster with ids different from the id of $A$. 

However, in our experiments, we observe that adding this triplet-centroid loss $L_{TC}$ does not improve the mAP or R1 for our application. We show this in Table \ref{table:centroid_vs_triplet_centroid}. 

We hypothesize that this is because our re-id application has much less data than the studies in \cite{do2019theoretically, yuan2020defense, zhang2020beyond, wang2019centroid, lagunes2020centroids, wieczorek2021unreasonable, alnissany2022modified}.
and because of the differences in the data distributions. 
Even though it uses centroids, the triplet-centroid loss is still calculated on a per-sample level with one triplet for each sample $A$. When there is sufficient amount of labelled data, this is not an issue since gradients are repeatedly calculated over multiple batches, thereby reducing any negative effect of individual outliers, and the network learns to average information over the samples.

For our limited data case, we obtain much better results by calculating a simple L2 loss using the centroids, instead of a triplet-centroid loss. Specifically, for each unique id $M_i$ in the batch, we partition the samples into two clusters. Cluster-I contains only the samples with the same id $M_i$, and cluster-II contains the remaining samples. We then calculate the embedding of the centroid of cluster-I and cluster-II. The loss function is simply the euclidean distance between these two centroids. We sum this loss for each unique id in the batch. This teaches the network to create embeddings that don't just push individual samples, but rather the clusters of different ids away from each other. Formally, for a batch with $K$ samples for each of $M$ ids, let $y(J)$ denote the id of the sample $J$ and $F_{J}$ the feature embedding for this sample $J$. Cluster-I for an id $M_i$ will have $K$ samples and cluster-II for this id will have $K \cdot M - K$ samples. The centroids of cluster-I and cluster-II for this id $M_i$ are calculated as follows:
\begin{align}
    C_{\text{cluster-I}}(M_i) &= \frac{1}{K} \cdot \sum_{J,  y(J) = M_i}(F_{J}) \\
    C_{\text{cluster-II}}(M_i) &= \frac{1}{K \cdot M - K} \cdot \sum_{J, y(J) \neq M_i}(F_{J})
\end{align}

The centroid loss for id $M_i$ is calculated as follows:

\begin{equation}
    L_{\text{Centroid}}(M_i) = ||C_{\text{cluster-I}}(M_i) - C_{\text{cluster-II}}(M_i)||^2
\end{equation}

The centroid loss for a batch is calculated as the sum over all ids in that batch.
\begin{equation}
\label{eq:centroid_loss}
    L_{\text{Centroid}} = \sum_{M_i} L_{\text{Centroid}}(M_i)
\end{equation}

While $L_{\text{Centroid}}$ is conceptually much simpler than the triplet-centroid loss $L_{TC}$, we show in Table \ref{table:centroid_vs_triplet_centroid}, that it does much better than $L_{TC}$ for our application.

The final loss function that we use is given by
\begin{equation}
\label{eq:loss}
    L = \alpha \cdot L_{T} + \beta \cdot L_{C} + \gamma \cdot L_{\text{Centroid}}
\end{equation}
where $\alpha$, $\beta$ and $\gamma$ are weights that are set empirically. $L_{T}$ and $L_{C}$ are the triplet and classification loss defined in Section \ref{sec:triplet_batchhard}.

\section{Experimental evaluation}
\label{sec:evaluation}
\subsection{Datasets}
\label{sec:datasets}

We use the ongoing SoccerNet Re-Identification Challenge 2022 dataset \cite{soccerNetv3, soccerNetv2} to evaluate our experiments. This is the first edition of this re-id challenge. This dataset is composed of 340993 players thumbnails extracted from the SoccerNet videos at different events, and images from their replays. The data is divided into train, validation, test and challenge splits. The training data has 248234 samples in total. 

The test split has 11777 query images and 34989 gallery images. However, the challenge website \cite{soccerNetv3} states that "player identity labels are derived from links between bounding boxes within an action and are therefore only valid within the given action. Consequently, player identity labels do not hold across actions and a given player has a different identity for each action he has been spotted in. For that reason, during the evaluation process, only samples within the same action are matched against each other." Therefore, for each query sample, we only need to compare against the gallery samples that have the same action. For evaluating on the test split, we train our networks on the train split only. The test split leaderboard is public and hence we can compare our performance with other methods.
The challenge split is composed of separate player thumbnails from different games and is sequestered, meaning that the ground-truth labels are unknown. There are 9021 query samples and 26082 gallery samples. Since the challenge is currently ongoing, the challenge split leaderboard is not public. Hence we do not know our position on the challenge split leaderboard. Therefore, we only report the performance of our network on the challenge split. We evaluate the same networks on the test and challenge splits. Only the train split is used for training in this case as well.

\subsection{Network}
\label{sec:network}
We conduct our evaluation using five different network architectures. Two of them are convolutional networks and the remaining three are vision transformers.

\textbf{Convolutional networks:} - The first is a ResNet50-fc512 network which is a ResNet50 \cite{he2016deep} with an extra fully connected layer of 512 output channels and a batchnorm layer as the penultimate layers, followed by a classification layer. We chose this network because the SoccerNet challenge provides it as a baseline. However, note that their baseline ResNet50-fc512 produces an mAP of 57.40 and a R1 of 45.89. We observed that by just adjusting the batchsize and a few hyperparameters, the same network produces an mAP of 70.3 and a R1 of 61.2 on the test split. So, we use the latter as our baseline. This network has 24.6M trainable parameters. The second convolutional network that we use is the OSNet\_x1\_0 \cite{zhou2019omni,zhou2021learning}. This network has produced compelling results on prior re-id literature and is very lightweight with 2.2M trainable parameters.

\textbf{Vision transformers:} The third network is a data-efficient image transformer DeiT-Tiny/16 \cite{touvron2021training} which is relatively lightweight with only 5.5M parameters but does well on image classification tasks. The fourth is a larger variant of this transformer called DeiT-S/16 \cite{touvron2021training} with 21.7M trainable parameters. The fifth network is the popular vision transformer ViT-B/16 \cite{dosovitskiy2020image} which produces SOTA results on image classification tasks. This network is quite huge with 57.7M trainable parameters. For all three transformer networks, we replace the final MLP classification layer with a fully connected layer with 512 output channels, a batchnorm layer followed by a final layer for classification. Hence all the five networks produce an embedding of length 512, for consistency. We choose these three transformers since the study in \cite{he2021transreid} shows good results for re-id applications with them.

\subsection{Training and hyperparameters}

We train the convolutional networks with ADAM \cite{kingma2014adam} optimizer and a linearly decaying learning rate scheduler. Transformers on the other hand are relatively trickier to train and therefore we use an ADAMW \cite{adamw} optimizer, with a linear warmup for the learning rate for the first 3000 iterations, followed by cosine annealing \cite{loshchilov2016sgdr}. We use the loss in equation \ref{eq:loss} with $\alpha = 0.9$, $\beta = 0.5$ and $\gamma = 0.5$ chosen empirically. We also carry out ablation studies where we disable the centroid loss, and use the triplet-centroid loss (equation \ref{eq:triplet_centroid}). The SoccerNet challenge provides the Torchreid \cite{zhou2019torchreid} library to develop our algorithms. Training is done on a single NVIDIA Ampere A100 GPU. To prevent selective/biased reporting of our improvements, each network is trained for 40 epochs with periodic checkpointing and we report the best mAP and R1 for each network. For fair comparison, the hyperparameters such as learning rate, weight decay etc. of each network are fixed across all experiments for that network.

\section{Results}
\label{sec:results}

\subsection{Evaluation on test split}

Table \ref{table:test_set_results} summarizes our results on the test split. We observe that for all five networks, hierarchical sampling when used with the additional centroid-loss (equation \ref{eq:centroid_loss}) increases the mAP by 7 - 11.5 and R1 by 8.8 - 14.9, when compared to random sampling and triplet loss.

\begin{table}
  \caption{Evaluation on the SoccerNet Re-Identification Challenge 2022 test split. Best numbers for each network are marked in bold font. The ViT-B/16 network with a mAP of 86.0 and a R1 of 81.5 is currently ranked \#1 on the test split leaderboard. Classification loss weight $\beta = 0.5$ for all cases.}
  \label{table:test_set_results}
  \centering
  \begin{tabular}{@{}lccclc@{}}
    \toprule
    Network & Sampling & Triplet loss & Centroid loss & mAP & R1 \\
    \midrule
    
    \multirow{4}{*}[-5pt]{ResNet50-fc512} & random & $\surd$ & $\times$ & 70.3 & 61.2\\
    \addlinespace[0.5em]
    & random & $\surd$ & $\surd$ & 75.4 & 68.7 \\
    \addlinespace[0.5em]
    & hierarchical & $\surd$ & $\times$ & 66.7 & 57.7\\
    \addlinespace[0.5em]
    & hierarchical & $\surd$ & $\surd$ & \textbf{81.8} & \textbf{76.1}\\
    \midrule

    \multirow{4}{*}[-5pt]{OSNet\_x1\_0} & random & $\surd$ & $\times$ & 76.4 & 69.2 \\
    \addlinespace[0.5em]
    & random & $\surd$ & $\surd$ & 78.5 & 72.6\\
    \addlinespace[0.5em]
    & hierarchical & $\surd$ & $\times$ & 75.8 & 69.1\\
    \addlinespace[0.5em]
    & hierarchical & $\surd$ & $\surd$ & \textbf{83.4} & \textbf{78.0}\\
    \midrule

    \multirow{4}{*}[-5pt]{DeiT-Tiny/16} & random & $\surd$ & $\times$ & 73.2 & 65.0\\
    \addlinespace[0.5em]
    & random & $\surd$ & $\surd$ & 74.9 & 67.1\\
    \addlinespace[0.5em]
    & hierarchical & $\surd$ & $\times$ & 80.5 & 74.4 \\
    \addlinespace[0.5em]
    & hierarchical & $\surd$ & $\surd$ & \textbf{82.2} & \textbf{76.2}\\
    \midrule
    
    \multirow{4}{*}[-5pt]{DeiT-S/16} & random & $\surd$ & $\times$ & 75.3 & 67.8\\
    \addlinespace[0.5em]
    & random & $\surd$ & $\surd$ & 78.9 & 72.3\\
    \addlinespace[0.5em]
    & hierarchical & $\surd$ & $\times$ & 81.0 & 74.7 \\
    \addlinespace[0.5em]
    & hierarchical & $\surd$ & $\surd$ & \textbf{84.3} & \textbf{79.4}\\
    \midrule

    \multirow{4}{*}[-5pt]{ViT-B/16} & random & $\surd$ & $\times$ & 75.7 & 68.2 \\
    \addlinespace[0.5em]
    & random & $\surd$ & $\surd$ & 78.4 & 72.0\\
    \addlinespace[0.5em]
    & hierarchical & $\surd$ & $\times$ & 81.4 & 75.4\\
    \addlinespace[0.5em]
    & hierarchical & $\surd$ & $\surd$ & \textbf{86.0} & \textbf{81.5}\\

    \bottomrule
  \end{tabular}
\end{table}

\subsection{Evaluation on challenge split}

Since the challenge is currently ongoing, the sequestered challenge split leaderboard is not public. Hence we do not know our position on the challenge split leaderboard. Also, since there is a hard limit on the number of submissions we can make to the challenge leaderboard, we only evaluate our best performing ViT-B/16 network trained with hierarchical sampling and the additional centroid loss on this split. It yields a mAP of 84.9 and a R1 of 80.1 on the challenge split.

\subsection{Ablation studies}

\subsubsection{Centroid loss vs triplet-centroid loss}
\label{sec:centroid_vs_triplet_centroid}

SOTA approaches such as the studies in \cite{do2019theoretically, yuan2020defense, zhang2020beyond, wang2019centroid, lagunes2020centroids, wieczorek2021unreasonable, alnissany2022modified} use triplet-centroid losses which is defined in equation \ref{eq:triplet_centroid}. As mentioned in Section \ref{sec:centroid_vs_triplet_centroid}, due to the limited training data per id, and the difference in data distributions, we observe that the triplet-centroid loss hardly improve mAP and R1. This is shown in Table \ref{table:centroid_vs_triplet_centroid}. Note that we only show a subset of results in this table due to space considerations. The full table is present in the supplementary material and our observations are similar across networks. The simple euclidean centroid loss is much more powerful for our task.

\begin{table}
  \caption{Comparison of centroid loss vs triplet-centroid loss. We only show a subset of results in this table due to space considerations. The full table is present in the supplementary material and our observations are similar across networks. Best numbers for each network are marked in bold font.}
  \label{table:centroid_vs_triplet_centroid}
  \centering
  \begin{tabular}{@{}lcccclc@{}}
    \toprule
    Network & Sampling & Triplet loss &  Centroid loss & Triplet-centroid loss & mAP & R1 \\
    \midrule
    


    
    \multirow{4}{*}[-5pt]{DeiT-Tiny/16} & \multirow{4}{*}[-5pt]{hierarchical} & $\surd$ & $\times$ & $\times$ & 80.5 & 74.4\\
    \addlinespace[0.5em]
    &  & $\surd$ & $\times$ & $\surd$ & 80.5 & 74.2\\
    \addlinespace[0.5em]
    &  & $\surd$ & $\surd$ & $\times$ & 82.2 & 76.2 \\
    \addlinespace[0.5em]
    &  & $\surd$ & $\surd$ &  $\surd$ & \textbf{82.3} & \textbf{76.4} \\
    \midrule

    


    \multirow{4}{*}[-5pt]{ViT-B/16} & \multirow{4}{*}[-5pt]{hierarchical} & $\surd$ & $\times$ & $\times$ & 81.4 & 75.4\\
    \addlinespace[0.5em]
    &  & $\surd$ & $\times$ & $\surd$ & 81.7 & 75.9\\
    \addlinespace[0.5em]
    &  & $\surd$ & $\surd$ & $\times$ &  \textbf{86.0} & \textbf{81.5}\\
    \addlinespace[0.5em]
    &  & $\surd$ & $\surd$ &  $\surd$ &  85.8 & 81.1\\

    \bottomrule
  \end{tabular}
\end{table}

\subsubsection{Using hierarchical sampling without the centroid loss}

Table \ref{table:test_set_results} also shows the mAP and R1 when the networks are trained without the centroid loss term. The results are quite interesting. For all three transformer networks, DeiT-Tiny/16, DeiT-S/16 and ViT-B/16, we see an increase of 5.7 - 7.3 in the mAP and an increase of 6.9 - 9.4 in the R1 by just using hierarchical sampling, without the centroid loss. However, for the two convolutional networks, we see a decrease. The mAP and R1 decrease by $\sim$3.6 for the ResNet50-fc512 which is, broadly speaking, the weakest of the five networks in terms of image classification accuracy. For the relatively more powerful OSNet\_x1\_0, the mAP and R1 decrease slightly by 0.6 and 0.1 respectively. We are quite surprised by this clear separation in the behavior of convolutional networks and transformers. More analysis is needed before we can draw any definitive conclusion from this. But we think it is interesting enough to point out.

\subsection{Additional results}

\begin{figure}[!ht]
     \begin{subfigure}{0.48\textwidth}
         \centering
         \includegraphics[width=1\textwidth,height=0.5\textwidth]{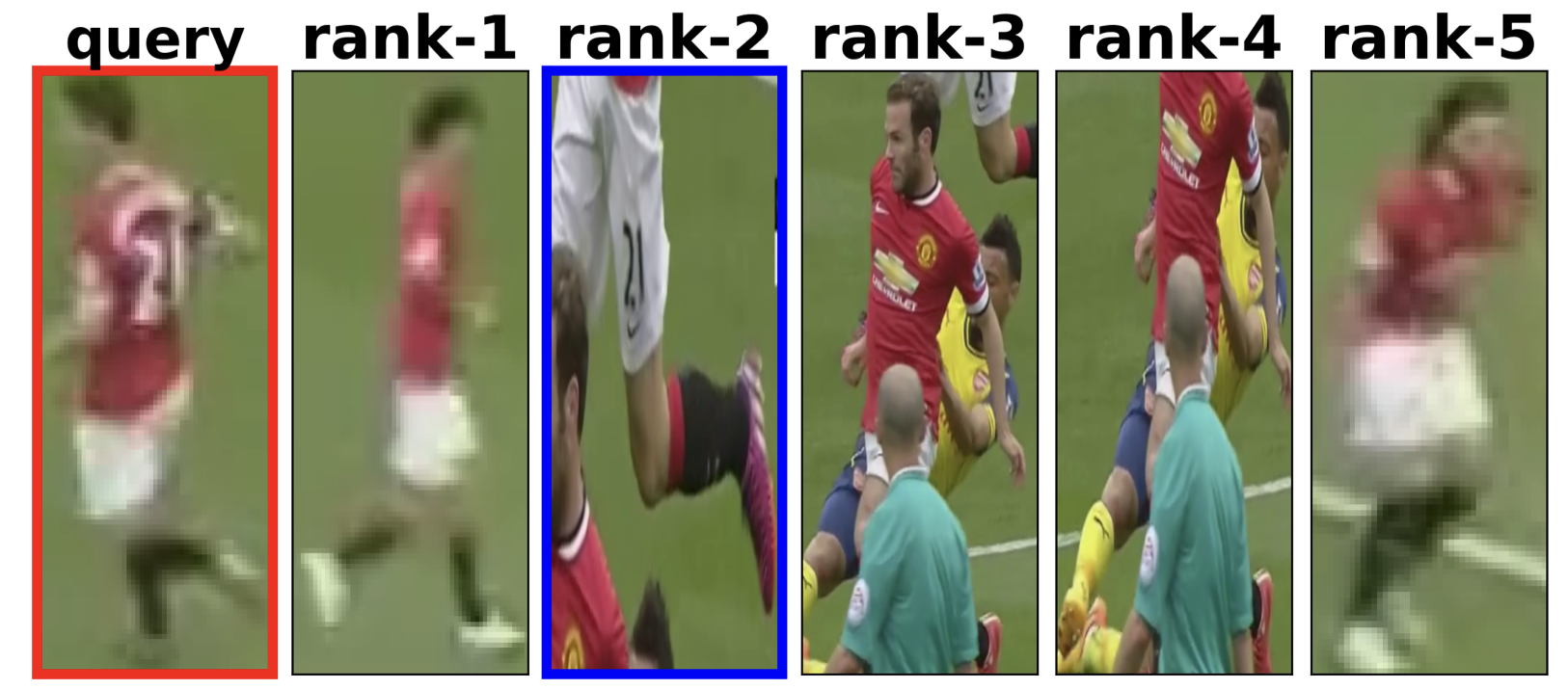}
         \caption{}
         \label{fig:add_a}
     \end{subfigure}
     \hfill
    \begin{subfigure}{0.48\textwidth}
        \centering
        \includegraphics[width=1\textwidth,height=0.5\textwidth]{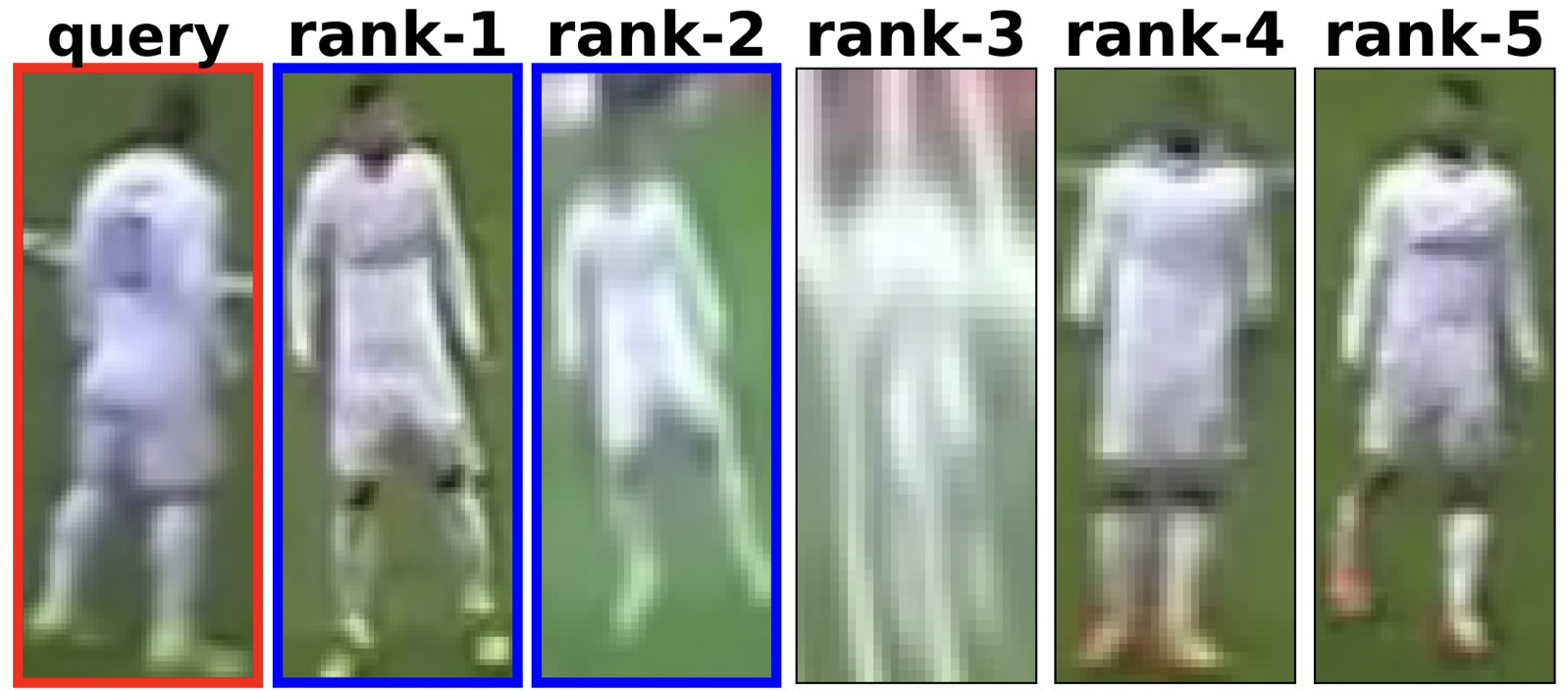}
        \caption{}
        \label{fig:add_b}
    \end{subfigure}
    \caption{Additional examples of successful re-id in challenging scenarios. The query is highlighted with a \textcolor{red}{red} border and the true matches in the corresponding gallery images are marked with a \textcolor{blue}{blue} border. Fig.~\ref{fig:add_a} shows that our procedure teaches the network to recognize the same number on the jersey in the query and in the shorts in the gallery. Fig.~\ref{fig:add_b} shows that the network learns to distinguish the subtle differences in the poses of the legs, despite all the players wearing the same white jersey. 
    }
    \label{fig:addfigures}
\end{figure}

Figs.~\ref{fig:results} and \ref{fig:addfigures} show examples of challenging cases where our procedure successfully re-identifies the player. The supplementary material contains examples of failure cases and a discussion of the limitations of our approach.



\section{Conclusions}
Re-identifying players in broadcast videos has a number of significant differences from surveillance re-id applications. These differences merit the needs for a specialized data sampling strategy and loss function. While we focus on soccer in this work due to availability of data, the ideas discussed in this work apply to many team sports. Hierarchical sampling can be easily extended to sport-specific cases, such as grouping by leagues, grouping by country etc. An unexpected result of our study is that a simple euclidean centroid loss is much more suited for our task with limited number of samples per id, when compared to SOTA triplet-centroid losses. Additional insights are included in the supplementary material.

\section{References}

\bibliography{sportsreid}

\begin{thebibliography}{34}
\providecommand{\natexlab}[1]{#1}
\providecommand{\url}[1]{\texttt{#1}}
\expandafter\ifx\csname urlstyle\endcsname\relax
  \providecommand{\doi}[1]{doi: #1}\else
  \providecommand{\doi}{doi: \begingroup \urlstyle{rm}\Url}\fi

\bibitem[soc()]{soccerNetv3}
Soccernet player re-identification challenge 2022.
\newblock \url{https://www.soccer-net.org/tasks/re-identification}.
\newblock Accessed: 2022-05-19.

\bibitem[Alnissany and Dayoub(2022)]{alnissany2022modified}
A.~Alnissany and Y.~Dayoub.
\newblock Modified centroid triplet loss for person re-identification.
\newblock 2022.

\bibitem[Bialkowski et~al.(2013)Bialkowski, Lucey, Wei, and
  Sridharan]{groupreid1}
A.~Bialkowski, P.~Lucey, X.~Wei, and S.~Sridharan.
\newblock Person re-identification using group information.
\newblock In \emph{2013 International Conference on Digital Image Computing:
  Techniques and Applications (DICTA)}, pages 1--6, 2013.
\newblock \doi{10.1109/DICTA.2013.6691512}.

\bibitem[Deli\'ege et~al.(2021)Deli\'ege, Cioppa, Giancola, Seikavandi,
  Dueholm, Nasrollahi, Ghanem, Moeslund, and Droogenbroeck]{soccerNetv2}
A.~Deli\'ege, A.~Cioppa, S.~Giancola, M.~J. Seikavandi, J.~V. Dueholm,
  K.~Nasrollahi, B.~Ghanem, T.~B. Moeslund, and M.~V. Droogenbroeck.
\newblock Soccernet-v2 : A dataset and benchmarks for holistic understanding of
  broadcast soccer videos.
\newblock In \emph{The IEEE Conference on Computer Vision and Pattern
  Recognition (CVPR) Workshops}, June 2021.

\bibitem[Do et~al.(2019)Do, Tran, Reid, Kumar, Hoang, and
  Carneiro]{do2019theoretically}
T.-T. Do, T.~Tran, I.~Reid, V.~Kumar, T.~Hoang, and G.~Carneiro.
\newblock A theoretically sound upper bound on the triplet loss for improving
  the efficiency of deep distance metric learning.
\newblock In \emph{Proceedings of the IEEE/CVF Conference on Computer Vision
  and Pattern Recognition}, pages 10404--10413, 2019.

\bibitem[Dosovitskiy et~al.(2020)Dosovitskiy, Beyer, Kolesnikov, Weissenborn,
  Zhai, Unterthiner, Dehghani, Minderer, Heigold, Gelly,
  et~al.]{dosovitskiy2020image}
A.~Dosovitskiy, L.~Beyer, A.~Kolesnikov, D.~Weissenborn, X.~Zhai,
  T.~Unterthiner, M.~Dehghani, M.~Minderer, G.~Heigold, S.~Gelly, et~al.
\newblock An image is worth 16x16 words: Transformers for image recognition at
  scale.
\newblock \emph{arXiv preprint arXiv:2010.11929}, 2020.

\bibitem[Hadi~Kiapour et~al.(2015)Hadi~Kiapour, Han, Lazebnik, Berg, and
  Berg]{Street2Shop}
M.~Hadi~Kiapour, X.~Han, S.~Lazebnik, A.~C. Berg, and T.~L. Berg.
\newblock Where to buy it: Matching street clothing photos in online shops.
\newblock In \emph{Proceedings of the IEEE international conference on computer
  vision}, pages 3343--3351, 2015.

\bibitem[He et~al.(2016)He, Zhang, Ren, and Sun]{he2016deep}
K.~He, X.~Zhang, S.~Ren, and J.~Sun.
\newblock Deep residual learning for image recognition.
\newblock In \emph{Proceedings of the IEEE conference on computer vision and
  pattern recognition}, pages 770--778, 2016.

\bibitem[He et~al.(2021)He, Luo, Wang, Wang, Li, and Jiang]{he2021transreid}
S.~He, H.~Luo, P.~Wang, F.~Wang, H.~Li, and W.~Jiang.
\newblock Transreid: Transformer-based object re-identification.
\newblock In \emph{Proceedings of the IEEE/CVF International Conference on
  Computer Vision}, pages 15013--15022, 2021.

\bibitem[Hermans et~al.(2017)Hermans, Beyer, and Leibe]{hermans2017defense}
A.~Hermans, L.~Beyer, and B.~Leibe.
\newblock In defense of the triplet loss for person re-identification.
\newblock \emph{arXiv preprint arXiv:1703.07737}, 2017.

\bibitem[Kingma and Ba(2014)]{kingma2014adam}
D.~P. Kingma and J.~Ba.
\newblock Adam: A method for stochastic optimization.
\newblock \emph{arXiv preprint arXiv:1412.6980}, 2014.

\bibitem[Lagunes-Fortiz et~al.(2020)Lagunes-Fortiz, Damen, and
  Mayol-Cuevas]{lagunes2020centroids}
M.~Lagunes-Fortiz, D.~Damen, and W.~Mayol-Cuevas.
\newblock Centroids triplet network and temporally-consistent embeddings for
  in-situ object recognition.
\newblock In \emph{2020 IEEE/RSJ International Conference on Intelligent Robots
  and Systems (IROS)}, pages 10796--10802. IEEE, 2020.

\bibitem[Li et~al.(2014)Li, Zhao, Xiao, and Wang]{CUHK03}
W.~Li, R.~Zhao, T.~Xiao, and X.~Wang.
\newblock Deepreid: Deep filter pairing neural network for person
  re-identification.
\newblock In \emph{Proceedings of the IEEE conference on computer vision and
  pattern recognition}, pages 152--159, 2014.

\bibitem[Loshchilov and Hutter(2016)]{loshchilov2016sgdr}
I.~Loshchilov and F.~Hutter.
\newblock Sgdr: Stochastic gradient descent with warm restarts.
\newblock \emph{arXiv preprint arXiv:1608.03983}, 2016.

\bibitem[Loshchilov and Hutter(2017)]{adamw}
I.~Loshchilov and F.~Hutter.
\newblock Decoupled weight decay regularization.
\newblock \emph{arXiv preprint arXiv:1711.05101}, 2017.

\bibitem[Matsukawa et~al.(2016)Matsukawa, Okabe, Suzuki, and Sato]{groupreid2}
T.~Matsukawa, T.~Okabe, E.~Suzuki, and Y.~Sato.
\newblock Hierarchical gaussian descriptor for person re-identification.
\newblock In \emph{Proceedings of the IEEE conference on computer vision and
  pattern recognition}, pages 1363--1372, 2016.

\bibitem[Penate-Sanchez et~al.(2020)Penate-Sanchez, Freire-Obregon,
  Lorenzo-Melian, Lorenzo-Navarro, and Castrillon-Santana]{penate2020tgc20reid}
A.~Penate-Sanchez, D.~Freire-Obregon, A.~Lorenzo-Melian, J.~Lorenzo-Navarro,
  and M.~Castrillon-Santana.
\newblock Tgc20reid: A dataset for sport event re-identification in the wild.
\newblock \emph{Pattern Recognition Letters}, 138:\penalty0 355--361, 2020.

\bibitem[Radh and Suresh()]{radhliterature}
V.~Radh and S.~Suresh.
\newblock A literature survey on person re-identification.

\bibitem[Touvron et~al.(2021)Touvron, Cord, Douze, Massa, Sablayrolles, and
  J{\'e}gou]{touvron2021training}
H.~Touvron, M.~Cord, M.~Douze, F.~Massa, A.~Sablayrolles, and H.~J{\'e}gou.
\newblock Training data-efficient image transformers \& distillation through
  attention.
\newblock In \emph{International Conference on Machine Learning}, pages
  10347--10357. PMLR, 2021.

\bibitem[Wang et~al.(2019{\natexlab{a}})Wang, Wang, Law, Rudzicz, and
  Brudno]{wang2019centroid}
J.~Wang, K.-C. Wang, M.~T. Law, F.~Rudzicz, and M.~Brudno.
\newblock Centroid-based deep metric learning for speaker recognition.
\newblock In \emph{ICASSP 2019-2019 IEEE International Conference on Acoustics,
  Speech and Signal Processing (ICASSP)}, pages 3652--3656. IEEE,
  2019{\natexlab{a}}.

\bibitem[Wang et~al.(2019{\natexlab{b}})Wang, Wang, Zheng, Wu, Zeng, and
  Satoh]{wang2019beyond}
Z.~Wang, Z.~Wang, Y.~Zheng, Y.~Wu, W.~Zeng, and S.~Satoh.
\newblock Beyond intra-modality: A survey of heterogeneous person
  re-identification.
\newblock \emph{arXiv preprint arXiv:1905.10048}, 2019{\natexlab{b}}.

\bibitem[Wang et~al.(2021)Wang, He, Tu, Zhao, Gao, Shen, and Feng]{groupreid5}
Z.~Wang, L.~He, X.~Tu, J.~Zhao, X.~Gao, S.~Shen, and J.~Feng.
\newblock Robust video-based person re-identification by hierarchical mining.
\newblock \emph{IEEE Transactions on Circuits and Systems for Video
  Technology}, 2021.

\bibitem[Wei et~al.(2018)Wei, Zhang, Gao, and Tian]{MSMT17}
L.~Wei, S.~Zhang, W.~Gao, and Q.~Tian.
\newblock Person transfer gan to bridge domain gap for person
  re-identification.
\newblock In \emph{Proceedings of the IEEE conference on computer vision and
  pattern recognition}, pages 79--88, 2018.

\bibitem[Wieczorek et~al.(2021)Wieczorek, Rychalska, and
  D{\k{a}}browski]{wieczorek2021unreasonable}
M.~Wieczorek, B.~Rychalska, and J.~D{\k{a}}browski.
\newblock On the unreasonable effectiveness of centroids in image retrieval.
\newblock In \emph{International Conference on Neural Information Processing},
  pages 212--223. Springer, 2021.

\bibitem[Wu et~al.(2019)Wu, Zheng, Zhang, Yuan, Cheng, Zhao, Lin, Zhao, Jiang,
  and Huang]{wu2019deep}
D.~Wu, S.-J. Zheng, X.-P. Zhang, C.-A. Yuan, F.~Cheng, Y.~Zhao, Y.-J. Lin,
  Z.-Q. Zhao, Y.-L. Jiang, and D.-S. Huang.
\newblock Deep learning-based methods for person re-identification: A
  comprehensive review.
\newblock \emph{Neurocomputing}, 337:\penalty0 354--371, 2019.

\bibitem[Ye et~al.(2018)Ye, Lan, Li, and Yuen]{groupreid3}
M.~Ye, X.~Lan, J.~Li, and P.~Yuen.
\newblock Hierarchical discriminative learning for visible thermal person
  re-identification.
\newblock In \emph{Proceedings of the AAAI Conference on Artificial
  Intelligence}, volume~32, 2018.

\bibitem[Yuan et~al.(2020)Yuan, Chen, Yang, and Wang]{yuan2020defense}
Y.~Yuan, W.~Chen, Y.~Yang, and Z.~Wang.
\newblock In defense of the triplet loss again: Learning robust person
  re-identification with fast approximated triplet loss and label distillation.
\newblock In \emph{Proceedings of the IEEE/CVF Conference on Computer Vision
  and Pattern Recognition Workshops}, pages 354--355, 2020.

\bibitem[Zeng et~al.(2020)Zeng, Ning, Wang, and Guo]{groupreid4}
K.~Zeng, M.~Ning, Y.~Wang, and Y.~Guo.
\newblock Hierarchical clustering with hard-batch triplet loss for person
  re-identification.
\newblock In \emph{Proceedings of the IEEE/CVF Conference on Computer Vision
  and Pattern Recognition}, pages 13657--13665, 2020.

\bibitem[Zhang et~al.(2020)Zhang, Lan, Zeng, Chen, and Chang]{zhang2020beyond}
Z.~Zhang, C.~Lan, W.~Zeng, Z.~Chen, and S.-F. Chang.
\newblock Beyond triplet loss: Meta prototypical n-tuple loss for person
  re-identification.
\newblock \emph{arXiv preprint arXiv:2006.04991}, 2020.

\bibitem[Zheng et~al.(2015)Zheng, Shen, Tian, Wang, Wang, and
  Tian]{Market-1501}
L.~Zheng, L.~Shen, L.~Tian, S.~Wang, J.~Wang, and Q.~Tian.
\newblock Scalable person re-identification: A benchmark.
\newblock In \emph{Proceedings of the IEEE international conference on computer
  vision}, pages 1116--1124, 2015.

\bibitem[Zheng et~al.(2021)Zheng, Tang, Teng, Ge, Liu, Qin, Qi, and
  Chen]{groupreid6}
Y.~Zheng, S.~Tang, G.~Teng, Y.~Ge, K.~Liu, J.~Qin, D.~Qi, and D.~Chen.
\newblock Online pseudo label generation by hierarchical cluster dynamics for
  adaptive person re-identification.
\newblock In \emph{Proceedings of the IEEE/CVF International Conference on
  Computer Vision}, pages 8371--8381, 2021.

\bibitem[Zhou and Xiang(2019)]{zhou2019torchreid}
K.~Zhou and T.~Xiang.
\newblock Torchreid: A library for deep learning person re-identification in
  pytorch.
\newblock \emph{arXiv preprint arXiv:1910.10093}, 2019.

\bibitem[Zhou et~al.(2019)Zhou, Yang, Cavallaro, and Xiang]{zhou2019omni}
K.~Zhou, Y.~Yang, A.~Cavallaro, and T.~Xiang.
\newblock Omni-scale feature learning for person re-identification.
\newblock In \emph{Proceedings of the IEEE/CVF International Conference on
  Computer Vision}, pages 3702--3712, 2019.

\bibitem[Zhou et~al.(2021)Zhou, Yang, Cavallaro, and Xiang]{zhou2021learning}
K.~Zhou, Y.~Yang, A.~Cavallaro, and T.~Xiang.
\newblock Learning generalisable omni-scale representations for person
  re-identification.
\newblock \emph{IEEE Transactions on Pattern Analysis and Machine
  Intelligence}, 2021.

\end{thebibliography}





\end{document}